\documentclass{article}

\PassOptionsToPackage{numbers, sort, compress}{natbib}

\usepackage[export]{adjustbox}
 \usepackage[preprint]{nips_2018}

\usepackage[utf8]{inputenc}
\usepackage{amsfonts}
\usepackage[T1]{fontenc}    
\usepackage{hyperref}       
\usepackage{url}            
\usepackage{booktabs}       
\usepackage{amsfonts}       
\usepackage{nicefrac}       
\usepackage{microtype}
\usepackage{graphicx}
\usepackage{amsmath,amssymb,amsthm}
\usepackage{subfig}

\newcommand{\overbar}[1]{\mkern 1.5mu\overline{\mkern-1.5mu#1\mkern-1.5mu}\mkern 1.5mu}
\title{Meta-Learning for Stochastic Gradient MCMC}

\author{
  Wenbo Gong$^{1*}$, Yingzhen Li$^{1*}$, Jos{\'e} Miguel Hern{\'a}ndez-Lobato$^1$ \\
  $^1$University of Cambridge\\
  \texttt{ \{wg242,yl494,jmh233\}@cam.ac.uk}
}

\begin{document}

\maketitle

\begin{abstract}
Stochastic gradient Markov chain Monte Carlo (SG-MCMC) has become increasingly popular for simulating posterior samples in large-scale Bayesian modeling. However, existing SG-MCMC schemes are not tailored to any specific probabilistic model, even a simple modification of the underlying dynamical system requires significant physical intuition.
This paper presents the first meta-learning algorithm that allows automated design for the underlying continuous dynamics of an SG-MCMC sampler. The learned sampler generalizes Hamiltonian dynamics with state-dependent drift and diffusion, enabling fast traversal and efficient exploration of neural network energy landscapes.
Experiments validate the proposed approach on both Bayesian fully connected neural network and Bayesian recurrent neural network tasks, showing that the learned sampler out-performs generic, hand-designed SG-MCMC algorithms, and generalizes to different datasets and larger architectures.
\end{abstract}

\section{Introduction}

There is a resurgence of research interests in Bayesian deep learning \citep{graves2011practical,blundell2015weight,hernandez2015probabilistic,hernandez2016black,gal2016dropout,ritter2018a}, which applies Bayesian inference to neural networks for better uncertainty estimation that is crucial for e.g.~better exploration in reinforcement learning \cite{deisenroth2011pilco,depeweg2017learning}, resisting adversarial attacks \citep{feinman2017detecting,li2017dropout,louizos2017multiplicative} and continual learning \citep{nguyen2018variational}. 
%
%
A popular approach to performing Bayesian inference on neural networks is stochastic gradient Markov chain Monte Carlo (SG-MCMC), which adds properly scaled Gaussian noise to a stochastic gradient ascent procedure \citep{welling2011bayesian}. Recent advances in this area further introduced optimization techniques such as pre-conditioning \citep{ahn2012bayesian,patterson2013stochastic}, annealing \citep{ding2014bayesian} and adaptive learning rates \citep{li2016preconditioned, chen2016bridging}. All these efforts have made SG-MCMC highly scalable to many deep learning tasks, including shape and texture modeling in computer vision \citep{li2016cvpr} and language modeling with recurrent neural networks \citep{gan2017scalable}.
However, inventing novel dynamics for SG-MCMC requires significant mathematical work to ensure the stationary distribution is the target distribution, which is less friendly to practitioners. Furthermore, many of these algorithms are designed as a generic sampling procedure, and the associated physical mechanism might not be best suited for sampling neural network weights. 

Can we automate the design of SG-MCMC algorithms which are tailored to the problem of sampling from certain types of densities, e.g. Bayesian neural network posterior distributions? This paper aims to answer this question by introducing \emph{meta-learning}, or \emph{learning to learn} techniques \citep{schmidhuber1987evolutionary,bengio1992optimization,naik1992meta,thrun1998learning}. The scope of meta-learning research is very broad, but the general idea is to train a \emph{learner} on one or multiple tasks in order to acquire common knowledge that generalizes to future tasks. Recent applications of meta-learning include learning to transfer knowledge to unseen few-shot learning tasks \citep{santoro2016meta,ravi2017fewshot,finn2017model}, and learning algorithms such as gradient descent \citep{andrychowicz2016learning,li2016learning,wichrowska2017learned}, Bayesian optimization \citep{chen2017learning} and reinforcement learning \citep{duan2016rl,wang2016learning}. Unfortunately these recent advances cannot be directly transfered to the world of MCMC samplers, since a naive neural network parameterization of the transition kernel does not guarantee the posterior distribution as a stationary distribution. 

We present to the best of our knowledge the first attempt towards meta-learning an SG-MCMC algorithm. 
%
%
Concretely, our contribution include:
\begin{itemize}
\setlength{\itemsep}{0.5pt}
\item An SG-MCMC sampler that extends Hamiltonian dynamics with \emph{learnable} diffusion and curl matrices. Once trained, the sampler can generalize to different datasets and architectures.  
\item Extensive evaluation of the proposed sampler on Bayesian fully connected neural networks and Bayesian recurrent neural networks, with comparisons to popular SG-MCMC schemes based on e.g.~Hamiltonian Monte Carlo \citep{chen2014stochastic} and pre-conditioned Langevin dynamics \citep{li2016preconditioned}.
\end{itemize}

\section{Background: a complete framework for SG-MCMC}
Consider sampling from a target density $\pi(\pmb{\theta})$ that is defined by an \emph{energy function}: $U(\pmb{\theta})$, $\pmb{\theta} \in \mathbb{R}^D$, $\pi(\pmb{\theta}) \propto \exp(-U(\pmb{\theta}))$. In this paper we focus on this sampling task in a Bayesian modeling set-up, i.e.~given observed data $\mathcal{D} = \{ \pmb{o}_n \}_{n=1}^N$, we define a probabilistic model $p(\mathcal{D}, \pmb{\theta}) = \prod_{n=1}^N p(\pmb{o}_n | \pmb{\theta})p(\pmb{\theta})$, and then the target density is the \emph{posterior distribution} $\pi(\pmb{\theta}) = p(\pmb{\theta} | \mathcal{D})$. 
Using Bayesian neural networks as an illustrating example, here $\pmb{o}_n = (\pmb{x}_n, \pmb{y}_n)$, and the model typically uses a Gaussian prior $p(\pmb{\theta}) = \mathcal{N}(\pmb{\theta}; \pmb{0}, \lambda^{-1}\pmb{I})$, and the energy function is defined as
\begin{equation}
U(\pmb{\theta}) = -\sum_{n=1}^N \log p(\pmb{y}_n | \pmb{x}_n, \pmb{\theta}) - \log p(\pmb{\theta}) = \sum_{n=1}^N \ell(\pmb{y}_n, \text{NN}_{\pmb{\theta}}(\pmb{x}_n) ) - \log p(\pmb{\theta}) ,
\label{eq:U}
\end{equation}
with $\ell(\pmb{y}, \hat{\pmb{y}})$ usually defined as the $\ell_2$ loss for regression or the cross-entropy loss for classification. A typical MCMC sampler constructs a Markov chain with a \emph{transition kernel}, and corrects the proposed samples with Metropolis-Hastings (MH) rejection steps. Some of these methods, e.g.~Hamiltonian Monte Carlo (HMC) \citep{duane1987hybrid, neal2011mcmc} and slice sampling \citep{neal2003slice}, further augment the state space with auxiliary variables $\pmb{r}$ and sample from the augmented distribution $\pi(\pmb{z})\propto \exp{(-H(\pmb{z}))}$, where $\pmb{z}=(\pmb{\theta},\pmb{r})$ and the corresponding Hamiltonian is $H(\pmb{z})=U(\pmb{\theta})+g(\pmb{r})$ such that $\int{\exp(-g(\pmb{r}))}d\pmb{r}=C$. Thus, marginalizing the auxiliary variable $\pmb{r}$ will not affect the stationary distribution $\pi(\pmb{\theta}) \propto \exp(-U(\pmb{\theta}))$.

For deep learning tasks, the observed dataset $\mathcal{D}$ often contains thousands, if not millions, of instances, making MH rejection steps computationally prohibitive. Fortunately this is mitigated by SG-MCMC, whose transition kernel is implicitly defined by a stochastic differential equation (SDEs) that leaves the target density invariant \citep{welling2011bayesian,ahn2012bayesian,patterson2013stochastic,chen2014stochastic, ding2014bayesian}. With carefully selected discretization step-size (like learning rates in optimization) the MH rejection steps are usually dropped. Also simulating one step of SG-MCMC only requires evaluating the gradient on a small mini-batch of data, which exhibits the same cost as many stochastic optimization algorithms. These two distinctive features make SG-MCMC highly scalable for sampling posterior distributions of neural network weights conditioned on big datasets.

Generally speaking, the continuous-time dynamics of an SG-MCMC method is governed by the following SDE (and the corresponding Markov process is called It\^{o} diffusion):\begin{equation}
d\pmb{z}=\pmb{f}(\pmb{z})dt+\sqrt{2\pmb{D}(\pmb{z})}d\pmb{W}(t),
\label{eq:Continuous SDE}
\end{equation}
with $\pmb{f}(\pmb{z})$ the deterministic drift, $\pmb{W}(t)$ the Wiener process, and $\pmb{D}(\pmb{z})$ the diffusion matrix. 
\citet{ma2015complete} derived an extensive framework of SG-MCMC samplers using advanced statistical mechanics \citep{yin2006existence,shi2012relation}, which directly parameterizes the drift term $\pmb{f}(\pmb{z})$ with the target density:
\begin{equation}
\pmb{f}(\pmb{z})=-[\pmb{D}(\pmb{z})+\pmb{Q}(\pmb{z})]\nabla{H(\pmb{z})}+\pmb{\Gamma}(\pmb{z}), \;\pmb{\Gamma}_i(\pmb{z})=\sum_{j=1}^d{\frac{\partial}{\partial \pmb{z}_j}(\pmb{D}_{ij}(\pmb{z})+\pmb{Q}_{ij}(\pmb{z}))}\,,
\label{eq:Gamma def}
\end{equation}
with $\pmb{Q}(\pmb{z})$ the curl matrix and $\pmb{\Gamma}(\pmb{z})$ a correction term. Remarkably \citet{ma2015complete} showed the completeness of their framework:
\begin{itemize}
\item[1.] $\pi(\pmb{z}) \propto \exp(-H(\pmb{z}))$ is a stationary distribution of the SDE (\ref{eq:Continuous SDE}) for any pair of positive semi-definite matrix $\pmb{D}(\pmb{z})$ and skew-symmetric matrix $\pmb{Q}(\pmb{z})$;
\item[2.] for any It\^{o} diffusion process that has the unique stationary distribution $\pi(\pmb{z})$, under mild conditions there exist $\pmb{D}(\pmb{z})$ and $\pmb{Q}(\pmb{z})$ matrices such that the process is governed by (\ref{eq:Continuous SDE}).
\end{itemize}
As a consequence, the construction of an SG-MCMC algorithm reduces to defining its $\pmb{D}$ and $\pmb{Q}$ matrices. Indeed \citet{ma2015complete} also casted existing SG-MCMC samplers within the framework, and proposed an improved version of SG-Riemannian-HMC. In general, an appropriate design of these two matrices leads to significant improvement on mixing as well as reduction of sample bias \citep{li2016preconditioned,ma2015complete}. However, historically this design has been based on strong physical intuitions from e.g.~Hamiltonian mechanics \citep{duane1987hybrid,neal2011mcmc} and thermodynamics \citep{ding2014bayesian}.  Therefore it can still be difficult for practitioners to understand and engineer the sampling method that best suited to their machine learning tasks.

%

%

\section{Meta-learning for SG-MCMC}

This section presents a meta-learning approach to learn an SG-MCMC sampler from data. Our aim is to design an appropriate parameterization of $\pmb{D}(\pmb{z})$ and $\pmb{Q}(\pmb{z})$, so that the sampler can be trained on small tasks with a meta-learning procedure, and generalize to more complicated densities in high dimensions. For simplicity, in this paper, we only augment the state-space by introducing one auxiliary variable $\pmb{p}$ called \emph{momentum} \citep{duane1987hybrid,neal2011mcmc}, although generalization to thermostat variable augmentation \citep{ding2014bayesian} is fairly straight-forward. Thus, the augmented state-space is $\pmb{z} = (\pmb{\theta}, \pmb{p})$ (i.e.~$\pmb{r}=\pmb{p}$), and the Hamiltonian is defined as $H(\pmb{z})=U(\pmb{\theta})+\frac{1}{2}\pmb{p}^T\pmb{p}$ (i.e.~with identity mass matrix for the momentum).

\subsection{Efficient parameterization of diffusion and curl matrices}

For neural networks, the dimensionality of $\pmb{\theta}$ can be at least tens of thousands. Thus, training and applying full $\pmb{D}(\pmb{z})$ and $\pmb{Q}(\pmb{z})$ matrices can cause huge computational burden, let alone gradient computations required by $\pmb{\Gamma}(\pmb{z})$. 
To address this, we define the preconditioning matrix as follows:
\begin{equation}
\begin{aligned}
\pmb{Q}(\pmb{z})=\left[\begin{array}{cc}
\pmb{0}&-\pmb{Q}_{f}(\pmb{z})\\
\pmb{Q}_f(\pmb{z})&\pmb{0}
\end{array}\right], \quad 
\pmb{D}(\pmb{z})=\left[\begin{array}{cc}
\pmb{0}&\pmb{0}\\
\pmb{0}&\pmb{D}_f(\pmb{z})
\end{array}
\right], \quad \quad \quad \\
\pmb{Q}_f(\pmb{z})=\text{diag}[\pmb{f}_{\phi_Q}(\pmb{z})], 
\quad
\pmb{D}_f(\pmb{z})=\text{diag}[\alpha \pmb{f}_{\phi_Q}(\pmb{z}) \odot \pmb{f}_{\phi_Q}(\pmb{z}) + \pmb{f}_{\phi_D}(\pmb{z}) + c], \quad \alpha, c >0,
\end{aligned}
\label{eq: Def of D and Q}
\end{equation}
where $\pmb{f}_{\theta_D}$ and $\pmb{f}_{\theta_Q}$ are neural network parameterized functions that will be detailed in section \ref{sec:network_design}, and $c$ is a small positive constant.
We choose $\pmb{D}_f$ and $\pmb{Q}_f$ to be diagonal for fast computation, although future work can explore low-rank matrix solutions. From \citet{ma2015complete}, our design has the \emph{unique} stationary distribution $\pi(\pmb{\theta})\propto\exp(-U(\pmb{\theta}))$ if $\pmb{f}_{\phi_D}$ is non-negative for all $\pmb{z}$. 

We discuss the role of each precondition matrix for better intuition. The curl matrix $\pmb{Q}(\pmb{z})$ in (\ref{eq:Continuous SDE}) mainly controls the deterministic drift forces introduced by the \textit{energy gradient} $\nabla_{\pmb{\theta}}{U(\pmb{\theta})}$ (as seen in many HMC-like procedures). Usually we only have the access to \textit{stochastic gradient} $\nabla_{\pmb{\theta}}{\tilde{U}(\pmb{\theta})}$ through data sub-sampling, so an additional friction term is needed to counter for the associated noise that mainly affects the dynamics of the momentum $\pmb{p}$. This explains the design of the diffusion matrix $\pmb{D}(\pmb{z})$ that uses $\pmb{D}_f(\pmb{z})$ to control the amount of friction and injected noise to the momentum.  
Furthermore $\pmb{D}_f(\pmb{z})$ should also account for the pre-conditioning effect introduced by $\pmb{Q}_f(\pmb{z})$, e.g, when the magnitude of $\pmb{Q}_f$ is large, without MH step we can only prevent divergence by increasing momentum friction. This explains the squared term $\pmb{f}_{\phi_Q}(\pmb{z}) \odot \pmb{f}_{\phi_Q}(\pmb{z})$ in $\pmb{D}_f$ design. The scaling positive constant $\alpha$ is heuristically selected following \citep{chen2014stochastic,ma2015complete} (see appendix).
Finally the extra term $\pmb{\Gamma}(\pmb{z}) = [\pmb{\Gamma}_{\pmb{\theta}}(\pmb{z}), \pmb{\Gamma}_{\pmb{p}}(\pmb{z})]$ is responsible for compensating the changes introduced by preconditioning matrices $\pmb{Q}(\pmb{z})$ and $\pmb{D}(\pmb{z})$.

The discretized dynamics of the state $\pmb{z} = (\pmb{\theta}, \pmb{p})$ with step-size $\eta$ and stochastic gradient $\nabla_{\pmb{\theta}}{\tilde{U}(\pmb{\theta})}$ is 
\begin{equation}
\begin{aligned}
\pmb{\theta}_{t+1}&=\pmb{\theta}_t+\eta\pmb{Q}_f(\pmb{z}_{t})\pmb{p}_t+\eta\pmb{\Gamma}_{\pmb{\theta}}(\pmb{z}_t),\\
\pmb{p}_{t+1}&=(1-\eta\pmb{D}_f(\pmb{z}_t))\pmb{p}_t-\eta\pmb{Q}_f(\pmb{z}_t)\nabla_{\pmb{\theta}}{\tilde{U}(\pmb{\theta})}+\eta\pmb{\Gamma}_{\pmb{p}}(\pmb{z}_t)+\pmb{\epsilon}, \quad \pmb{\epsilon} \sim \mathcal{N}(\pmb{0}, 2\eta \pmb{D}_f(\pmb{z})).
\end{aligned}
\label{eq: update rule}
\end{equation}
%
Again we notice that $\pmb{Q}_{f}$ is responsible for the acceleration of $\pmb{\theta}$, and from the $(1-\eta\pmb{D}_{f})$ term in the update equation of $\pmb{p}$, we see that $\pmb{D}_f$ controls the friction introduced to the momentum.
Observing that the noisy gradient is approximately Gaussian distributed in the big data setting, \citet{ma2015complete} further suggested a correction scheme to counter for stochastic gradient noise, which samples the Gaussian noise $\pmb{\epsilon} \sim \mathcal{N}(\pmb{0}, 2\eta \pmb{D}_f(\pmb{z}) - \eta^2 \tilde{B}(\pmb{\theta}))$ with an empirical estimate of the gradient variance $\tilde{B}(\pmb{\theta})$ instead. These corrections can be dropped when the discretization step-size $\eta$ is small, therefore we do not consider them in our experiments. 

\subsection{Choices of inputs to the neural networks}
\label{sec:network_design}

We now present detailed functional forms for $\pmb{f}_{\phi_Q}$ and $\pmb{f}_{\phi_D}$. When designing these, our goal was to achieve a good balance between generalization power and computational efficiency. 
%
%
%
Recall that the curl matrix $\pmb{Q}$ mainly controls the drift of the dynamics, and the desired behavior is that it should produce accelerations for fast traverse through low density regions. One useful source of information to identify these regions is the energy function $U(\pmb{\pmb{\theta}})$ which can be used to determine if the particles have reached high density regions.\footnote{The energy gradient $\nabla_{\pmb{\theta}}U(\pmb{\theta})$ is also informative here, however, it requires computing the diagonal Hessian for $\pmb{\Gamma}(\pmb{z})$ which is costly for high dimension problems. For similar reasons we do not consider (diagonal) Fisher information matrix or Hessian as an input of $\pmb{f}_{\phi_Q}$.} 
%
We also include the momentum $p_i$ to the inputs of $\pmb{f}_{\phi_Q}$, allowing the $\pmb{Q}$ matrix to observe the velocity information of the particles.
We further add an offset $\beta$ to $\pmb{Q}$ to prevent the vanishing of this matrix. Putting all of them together, we define the $i^{\text{th}}$ element of $\pmb{f}_{\phi_Q}$ as 
\begin{equation}
\pmb{f}_{\phi_Q,i}(\pmb{z})= \beta +f_{\phi_Q}(U(\pmb{\theta}),p_i) 
\label{eq:Q param}
\end{equation}

The corresponding $\pmb{\Gamma}$ term requires both $\partial_{\theta_i}f_{\phi_Q}(U(\pmb{\theta}),p_i)$ and $\partial_{p_i}f_{\phi_Q}(U(\pmb{\theta}),p_i)$. The energy gradient $\partial_{\theta_i}U(\pmb{\theta})$ also appears in (\ref{eq: update rule}) so it remains to compute $\partial_U \pmb{f}_{\phi_Q}$, which, along with $\partial_{p_i}f_{\phi_Q}(U(\pmb{\theta}),p_i)$, can be obtained by automatic differentiation \citep{abadi2016tensorflow}.

Matrix $\pmb{D}$ is responsible for the friction term and the stochastic gradient noise, which are crucial for better exploration around high density regions. Therefore we also add energy gradient $\partial_{\theta_i}U(\pmb{\theta})$ to the inputs, meaning that the $i^{\text{th}}$ element of $\pmb{f}_{\phi_D}$ is
\begin{equation}
\pmb{f}_{\phi_D,i}(\pmb{z})=f_{\phi_D}(U(\pmb{\theta}),p_i,\partial_{\theta_i}U(\pmb{\theta}))
\label{eq:D param}
\end{equation}
By construction of the $\pmb{D}$ matrix, the $\pmb{\Gamma}$ vector only requires $\nabla_{\pmb{p}} \pmb{D}_f$, therefore the Hessian of the energy function is not required. 

In practice both $U(\pmb{\theta})$ and $\partial_{\theta_i}U(\pmb{\theta})$ are replaced by their stochastic estimates $\tilde{U}(\pmb{\theta})$ and $\partial_{\theta_i}\tilde{U}(\pmb{\theta})$, respectively.
To keep the scale of the inputs roughly the same across tasks, we rescale all the inputs using statistics computed by simulating the sampler with random initialized $\pmb{f}_{\phi_D}$ and $\pmb{f}_{\phi_Q}$.  
When the computational budget is limited, we replace the exact gradient computation required by $\pmb{\Gamma}(\pmb{z})$ with finite difference approximations. We refer the reader to the appendix for details.

\subsection{Loss function design for meta-learning}
Another challenge is to design a meta-learning procedure for the sampler to encourage faster convergence and low bias on test tasks. 
%
%
To achieve these goals we propose two loss functions that we named as \emph{cross-chain loss} and \emph{in-chain loss}. From now on we consider the discretized dynamics and define $q_t(\pmb{\theta}|\mathcal{D})$ as the marginal distribution of the random variable $\pmb{\theta}$ at time $t$. 

\paragraph{Cross-chain loss}
We introduce \emph{cross-chain loss} that encourages the sampler to exhibit fast convergence. Since the framework guarantees the sampler to have the target density $\pi(\pmb{\theta}) \propto \exp(-U(\pmb{\theta}))$ as the unique stationary distribution, fast convergence means that $\mathrm{KL}[q_t||\pi]$ is close to zero when $t$ is small. Therefore this KL-divergence becomes a sensible objective to minimize, which is equivalent to maximizing the variational lower-bound (or ELBO): $\mathcal{L}_{\text{VI}}^t(q_t) = -\mathbb{E}_{q_t}[U(\pmb{\theta})] + \mathbb{H}[q_t]$ \citep{jordan1999introduction,beal2003variational}.
We further make the objective doubly stochastic: (1) the energy term is further approximated by its stochastic estimates $\tilde{U}(\pmb{\theta})$; (2) we use Monte Carlo variational inference (MCVI) \citep{ranganath2014black,blundell2015weight} which estimates the lower-bound with samples $\pmb{\theta}^t_k \sim q_t(\pmb{\theta}_t|\mathcal{D}), k = 1, ..., K$. These particles $\{\pmb{\theta}_k^t \}_{k=1, t=1}^{K, T}$ are obtained by simulating $K$ parallel Markov chains with the sampler, and the cross-chain loss is defined by accumulating the lower-bounds through time: 
\begin{equation}
\mathcal{L}_{\text{cross-chain}}=\frac{1}{T}\sum_{t=1}^T{\mathcal{L}_{\text{VI}}^t(\{\pmb{\theta}^t_k\}_{k=1}^K)}, \quad \mathcal{L}_{\text{VI}}^t(\{\pmb{\theta}^t_k\}_{k=1}^K)= -\frac{1}{K}\sum_{k=1}^K \left[ \tilde{U}(\pmb{\theta}_k^t) + \log q_t(\pmb{\theta}_k^t|\mathcal{D}) \right].
\end{equation}
By minimizing this objective, we can improve the convergence of the sampler, especially at the early times of the Markov chain. The objective also takes the sampler bias into account because the two distributions will match when the KL-divergence is minimized. 


\paragraph{In-chain loss}
For very big neural networks, simulating multiple Markov chains is prohibitively expensive. The issue is mitigated by \emph{thinning}, which collects samples for every $\tau$ step (after burn-in). Effectively, after thinning the samples are drawn from the averaged distribution $\bar{q}(\pmb{\theta}|\mathcal{D}) = \frac{1}{ \lfloor T / \tau \rfloor} \sum_{s=1}^{\lfloor T / \tau \rfloor} q_{s\tau}(\pmb{\theta})$. The in-chain loss is therefore defined as the ELBO evaluated at the averaged distribution $\bar{q}$, which is then approximated by Monte Carlo with samples $\pmb{\Theta}_k^{T, \tau} = \{\pmb{\theta}_k^{s\tau} \}_{s=1}^{\lfloor T / \tau \rfloor}$ obtained by thinning:
\begin{equation}
\mathcal{L}_{\text{in-chain}} = \frac{1}{K} \sum_{k=1}^K \mathcal{L}_{\text{VI}}^k \left( \pmb{\Theta}_k^{T, \tau} \right),
\quad
\mathcal{L}_{\text{VI}}^k \left( \pmb{\Theta}_k^{T, \tau} \right) = -\frac{1}{\lfloor T / \tau \rfloor}\sum_{s=1}^{\lfloor T / \tau \rfloor} \left[ \tilde{U}(\pmb{\theta}_k^{s\tau}) + \log \bar{q}(\pmb{\theta}_k^{s\tau}|\mathcal{D}) \right].
\end{equation}

\paragraph{Gradient approximation} Gradient-based optimization requires the gradient $\nabla_{\phi} \log q_t(\pmb{\theta})$ for cross-chain loss and $\nabla_{\phi} \log \bar{q}(\pmb{\theta})$ for in-chain loss. Since the density of $q_t$ is intractable, we leverage the recently proposed Stein gradient estimator \citep{li2017stein} for gradient approximation. Precisely, by the chain rule we have $\nabla_{\phi} \log q_t(\pmb{\theta}) = \nabla_{\phi} \pmb{\theta} \nabla_{\pmb{\theta}} \log q_t(\pmb{\theta})$. 
Denote $\pmb{G} =(\nabla_{\pmb{\theta}_1^t}\log q_t(\pmb{\theta}_1^t)),\ldots,\nabla_{\pmb{\theta}_K^t}\log q_t(\pmb{\theta}_K^t))^T$ as the matrix collecting the gradients of $\log q_t(\pmb{\theta})$ at the sampled locations $\{ \pmb{\theta}_k^t \}_{k=1}^K \sim q_t$. The recipe first constructs a kernel matrix $\pmb{K}$ with $\pmb{K}_{ij} = \mathcal{K}(\pmb{\theta}^t_i, \pmb{\theta}^t_j)$, then compute an estimate of the $\pmb{G}$ matrix by 
$
\pmb{G} \approx -(\pmb{K}+\lambda\pmb{I})^{-1}\langle\nabla,\pmb{K}\rangle
$, where $
\langle\nabla,\pmb{K}\rangle_{ij} = \sum_{k=1}^K \partial_{\pmb{\theta}^t_{k, j}} \mathcal{K}(\pmb{\theta}^t_k, \pmb{\theta}^t_i)$. In our experiments, we use RBF kernels, and the corresponding gradient estimator has simple analytic form that can be computed efficiently in $\mathcal{O}(K^2D + K^3)$ time (usually $K \ll D$).

\section{Related work}

Since the development of stochastic gradient Langevin dynamics \citep[SGLD,][]{welling2011bayesian}, SG-MCMC has been increasingly popular for sampling from posterior densities of big models conditioned on big data. In detail, \citet{chen2014stochastic} scaled up HMC with stochastic gradients, \citet{ding2014bayesian} further augmented the state space with a temperature auxiliary variable and performed sampling in the joint space and \citet{SpringenbergKFH16} improved robustness through scale adaptation. The SG-MCMC extensions to Riemannian Langevin dynamics and HMC \citep{girolami2011riemann} have also been proposed \citep{patterson2013stochastic, ma2015complete}.
Our proposed sampler architecture further generalizes SG-Riemannian-HMC as it decouples the design of $\pmb{D}(\pmb{z})$ and $\pmb{Q}(\pmb{z})$ matrices, and the detailed functional form of these two matrices are also learned from data.

Our approach is closely related to the recent line of work on learning optimization algorithms. Specifically, \citet{andrychowicz2016learning} trained a recurrent neural network (RNN) based optimizer that transfers to similar tasks with supervised learning. Later \citet{chen2017learning} generalized this approach to Bayesian optimization \citep{brochu2010tutorial,snoek2012practical} which is gradient-free. 
We do not use RNN in our approach as it cannot be represented within the framework of \citet{ma2015complete}. We leave the combination of learnable RNN proposals to future work.
Also \citet{li2017stein} presented an initial attempt to meta-learn an approximate inference algorithm, which simply combined the stochastic gradient and the Gaussian noise with a neural network. Thus the stationary distribution of that sampler (if it exists) is only an approximation to the exact posterior. On the other hand, the proposed sampler (with $\eta \rightarrow 0$) is guaranteed to be correct by complete framework \cite{ma2015complete}.
Very recently \citet{wu2018understanding} discussed short-horizon meta-objectives for learning optimizers can cause a serious issue for long-time generalization. We found this bias is less severe in our approach, again due to the fact that the learned sampler is provably correct.

Recent research also considered improving HMC with a trainable transition kernel. \citet{salimans2015markov} improved upon vanilla HMC by introducing a trainable re-sampling distribution for the momentum. 
\citet{song2017nice} parameterized the HMC transition kernel with a trainable invertible transformation called non-linear independent components estimation (NICE) \citep{dinh2014nice}, and learned this operator with Wasserstein adversarial training \citep{arjovsky2017wasserstein}.
\citet{levy2017generalizing} generalized HMC by augmenting the state space with a binary direction variable, and they parameterized the transition kernel with a non-volume preserving invertible transformation that is inspired by real-valued non-volume preserving (RealNVP) flows \citep{dinh2016density}. The sampler is then trained with expected squared jump distance \citep{pasarica2010adaptively}.
We do not explore the adversarial training idea in this paper as for very high dimensional distributions these techniques become less reliable.
Also the jump distance does not explicitly take the sampling bias and convergence speed into account.
More importantly, the purpose of these approaches is to directly improve the HMC-like sampler on the \emph{target distribution}, and with NICE/RealNVP parametrization it is difficult to generalize the sampler to densities of different dimensions. In contrast, our goal is to learn an SG-MCMC sampler that can later be transferred to sample from \emph{different} Bayesian neural network posterior distributions, which will typically have \emph{different} dimensionality and include tens of thousands of random variables.

\section{Experiments}
We evaluate the meta-learned SG-MCMC sampler, which is referred to as NNSGHMC or the meta sampler in the following. Detailed test set-ups are reported in the appendix. Code is available at \url{https://github.com/WenboGong/MetaSGMCMC}.

\subsection{Synthetic example: sampling Gaussian random variables with noisy gradients}
\begin{figure}
\centering
\includegraphics[width=1\textwidth,height=0.33\textwidth]{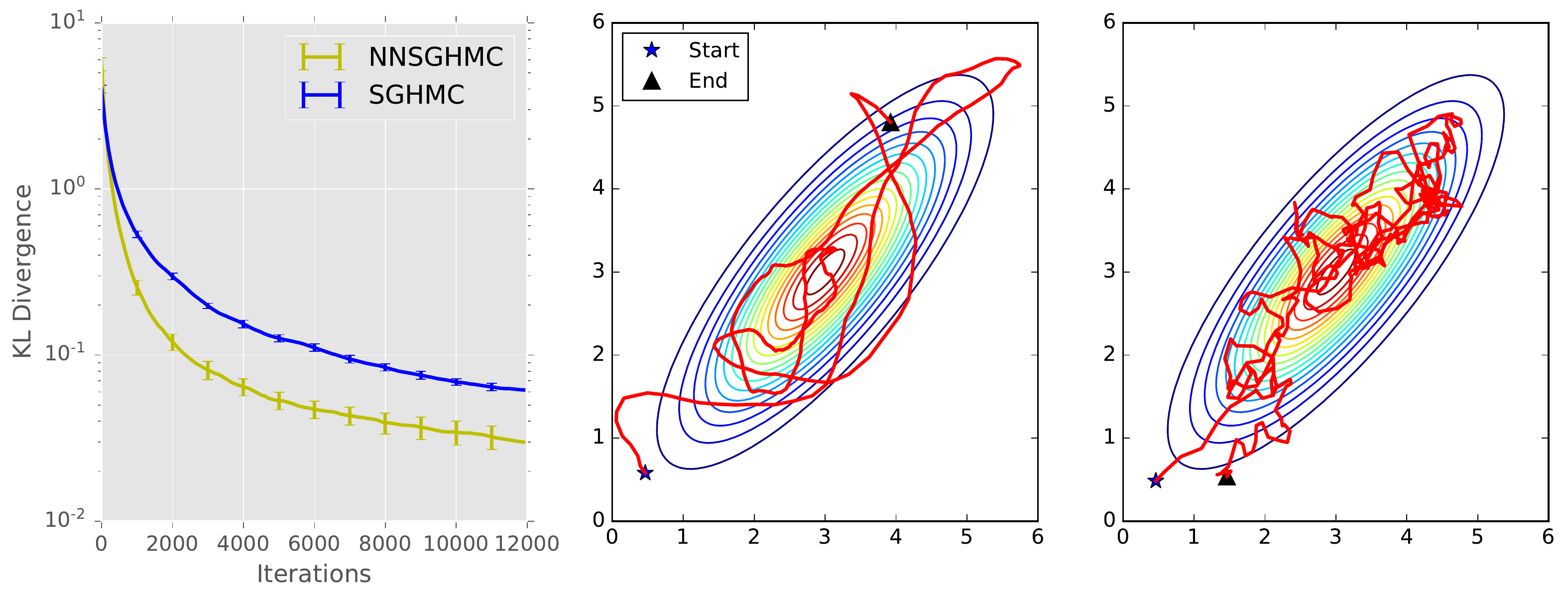}
\caption{(Left) Sampler's bias measured by KL.  (Middle) NNSGHMC trajectory plot on a 2D-Gaussian with manually injected gradient noise. (Right) SGHMC plot for the same settings.}
\label{fig:Toy}
\vspace{-0.1in}
\end{figure}

We first consider sampling Gaussian variables to demonstrate fast convergence and low bias of the meta sampler. To mimic stochastic gradient settings, we manually inject Gaussian noise with unit variance to the gradient as suggested by \citep{chen2014stochastic}. 
The training density is a 10D Gaussian with randomly generated diagonal covariance, and the test density is a 20D Gaussian. 
For evaluation, we simulate $K=50$ parallel chains for $T=12000$ steps. Then we follow \cite{ma2015complete} to evaluate the sampler's bias is measured by the KL divergence from the empirical Gaussian estimate to the ground truth.  
Results are visualized on the left panel of Figure \ref{fig:Toy}, showing that the meta sampler both converges much faster and achieves lower bias compared to SGHMC. The effective sample size\footnote{Implementation follows the ESS function in the BEAST package \url{http://beast.community}.} for SGHMC and NNSGHMC are \textbf{22} and \textbf{59}, respectively, again indicating better efficiency of the meta sampler. For illustration purposes, we also plot in the other two panels the trajectory of a particle by simulating NNSGHMC (middle) and SGHMC (right) on a 2D Gaussian for fixed amount of time $\eta T$. This confirms that the meta sampler explores more efficiently and is less affected by the injected noise. 


\subsection{Bayesian feedforward neural network}
Next we consider Bayesian neural network classification on MNIST data with three generalization tests: \emph{network architecture generalization} (NT), \emph{activation function generalization} (AF) and \emph{dataset generalization} (Data). In all tests the sampler is trained with a 1-hidden layer MLP (20 units, ReLU activation) as the underlying model for the target distribution $\pi(\pmb{\theta})$. We also report long-time horizon generalization results, meaning that the simulation time steps in test time is much longer than that of training \citep[cf.][]{andrychowicz2016learning}. 
Algorithms in comparison include SGLD \cite{welling2011bayesian}, SGHMC \cite{chen2014stochastic} and preconditioned SGLD \citep[PSGLD,][]{li2016preconditioned}. Note that PSGLD uses RMSprop-like preconditioning techniques \citep{tieleman2012rmsprop} that requires moving average estimates of the gradient's second moments. Therefore the underlying dynamics of PSGLD cannot be represented within our framework (\ref{eq: Def of D and Q}). Thus we main focus on comparisons with SGLD and SGHMC, and leave the PSGLD results as reference. The discretization step-sizes for the samplers are tuned on the validation dataset for each task.

\begin{figure}[t]
\centering
\includegraphics[width=1.\textwidth,height=0.35\textwidth]{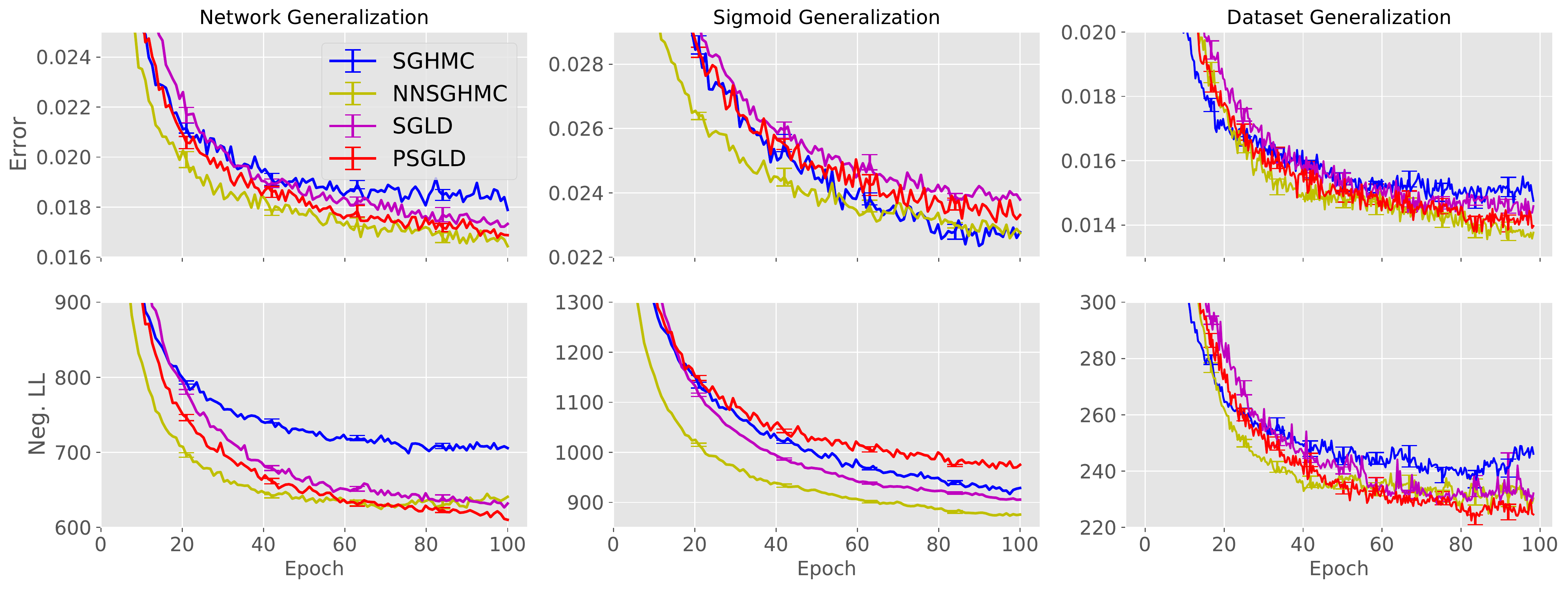}
\caption{Learning curves on test error (top) and negative test LL (bottom).}
\label{fig: Error MNIST}

\end{figure}

\paragraph{Architecture generalization (NT)}
In this test we use the trained sampler to draw samples from the posterior distribution of a \emph{2-hidden layer} MLP with \emph{40 units} and ReLU activations. 
%
%
Figure \ref{fig: Error MNIST} shows learning curves of test error and negative test log-likelihood (NLL) for 100 epochs, where the final performance is reported in Table \ref{tab: NetworkGen Result}. Overall NNSGHMC achieves the fastest convergence even when compared with PSGLD. It has the lowest test error compared to SGLD and SGHMC. NNSGHMC's final test LL is on par with SGLD and slightly worse than PSGLD, but it is still better than SGHMC. 

\begin{figure}[!t]
\centering
\includegraphics[width=1\textwidth,height=0.28\textwidth]{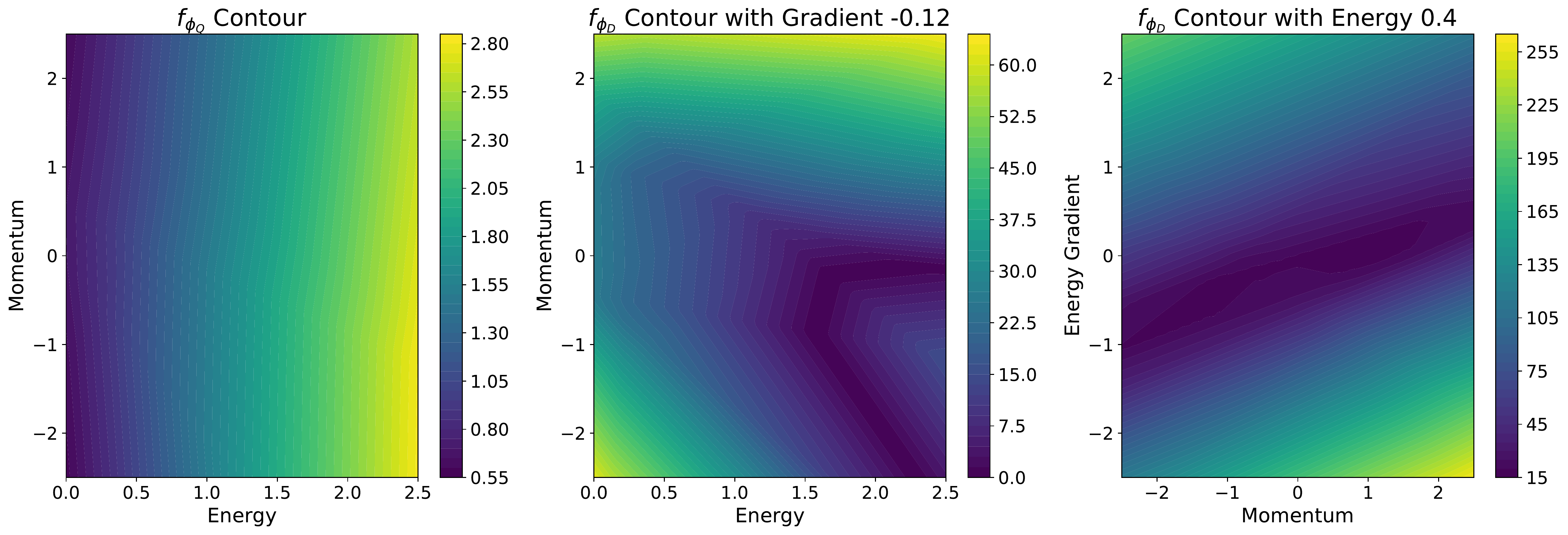}
\caption{(Left) The contour plot of function $\pmb{f}_{\phi_Q}$ (Middle) The contour plot for $\pmb{f}_{\phi_D}$ for dimension 1 and 2 with fixed $-\nabla_{\pmb{\theta}}{U(\pmb{\theta})}$ (Right) The same plot for $\pmb{f}_{\phi_D}$ for dimension 2 and 3 with fixed energy.}
\label{fig: D Q Strategy}
\vspace{-0.1in}
\end{figure}
\begin{table}[h]
\centering
\caption{The final performance for the samplers, averaged over 10 independent runs.}
\resizebox{\columnwidth}{!}{
\begin{tabular}{ccccccc}
\hline
Sampler& NT Err.&AF Err & Data Err& NT NLL &AF NLL & Data NLL\\
\hline
NNSGHMC & \textbf{98.36$\pm \pmb{0.02}$}\% & \textbf{97.72$\pm \pmb{0.02}$}\%& \textbf{98.62$\pm \pmb{0.02}$}\%&640$\pm 6.25$&\textbf{875$\pm\pmb{3.19}$}&230$\pm 3.23$\\
SGHMC & 98.21$\pm0.01$\% & \textbf{97.72$\pm \pmb{0.01}$}\% &98.52$\pm 0.03$\% &705$\pm 3.44$&929$\pm 2.95$&246$\pm 5.43$\\
SGLD & 98.27$\pm 0.02$\% & 97.62$\pm 0.02$\% &98.54$\pm 0.01$\% &631$\pm 3.15$ &905$\pm 2.36$&232$\pm 1.93$\\
PSGLD &98.31$\pm 0.02$\% & 97.67$\pm 0.02$\% &98.60$\pm 0.02$\% &\textbf{610$\pm\pmb{2.93}$}&975$\pm 4.41$&\textbf{224$\pm\pmb{1.97}$}\\
\hline
\end{tabular}
}
\label{tab: NetworkGen Result}
\vspace{-0.1in}
\end{table}

\paragraph{Activation function generalization (AF)}
Next we replace the test network's activation with \textbf{sigmoid} function and re-run the same test as before.
%
Again results in Figure \ref{fig: Error MNIST} and Table \ref{tab: NetworkGen Result} show that NNSGHMC converges faster than others for both test error and NLL. It also achieves the best NLL results among all samplers, and the same test error as SGHMC. 

\paragraph{Dataset generalization (Data)}
In this test we split MNIST into \emph{training task} (classifying digits 0-4) and \emph{test task} (digits 5-9). We train the meta sampler on the training task using the small MLP as before, and evaluate the learned sampler on the test task with the larger MLP. From the plots, we see that NNSGHMC, although a bit slower at start, catches up quickly and proceeds to lower error. The difference between these samplers NLL results is marginal, and NNSGHMC is on par with PSGLD.

\paragraph{Learned strategies}
For better intuition, we visualize in Figure \ref{fig: D Q Strategy} the contours of $\pmb{f}_{\phi_D}$ and $\pmb{f}_{\phi_Q}$ as a function of their inputs. Recall that the curl matrix $\pmb{Q}$ is determined by $\pmb{f}_{\phi_Q}$.
From the left panel, $\pmb{f}_{\phi_Q}$ has learned a nearly linear strategy w.r.t.~the energy and small variations w.r.t. the momentum. This enables the sampler for fast traversal through low density (high energy) regions and better exploration at high density (low energy) area.

The strategy learned for the diffusion matrix $\pmb{D}$ is rather interesting. Recall that $\pmb{D}$ is parametrized by both $\pmb{f}_{\phi_D}$ and $\pmb{f}_{\phi_Q}\odot\pmb{f}_{\phi_Q}$ (Eq.\ref{eq: Def of D and Q}). 
Since Figure \ref{fig: D Q Strategy} (left) indicates that $\pmb{f}_{\phi_Q}$ is large in high energy regions, the amount of friction is adequate, so $\pmb{f}_{\phi_D}$ tends to reduce its output to maintain momentum (see the middle plot). By contrast, at low energy regions $\pmb{f}_{\phi_D}$ increases to add friction in order to prevent divergence. 
The right panel visualizes the interactions between the momentum and the mean gradient $-\frac{1}{N}\nabla_{\pmb{\theta}}U(\pmb{\theta})$ at a fixed energy level. This indicates that the meta sampler has learned a strategy to prevent overshoot by producing large friction, indeed $\pmb{f}_{\phi_D}$ returns large values when the signs of the momentum and the gradient differ.

\subsection{Bayesian recurrent neural networks}
Lastly we consider a more challenging setup: sequence modeling with Bayesian RNNs. Here a single datum is a sequence $\pmb{o}_n = \{ \pmb{x}_n^1, ..., \pmb{x}_n^T \}$ and the log-likelihood is defined as $\log p(\pmb{o}_n | \pmb{\theta}) = \sum_{t=1}^T \log p(\pmb{x}_t^n|\pmb{x}_1^n,\ldots,\pmb{x}_{t-1}^n,\pmb{\theta})$, with each of the conditional densities produced by a gated recurrent unit (GRU) network \citep{cho2014learning}. We consider four polyphonic music datasets for this task: Piano-midi (Piano) as training data, and Nottingham (Nott), MuseData (Muse) and JSB chorales (JSB) for evaluation. 
The meta sampler is trained on a small GRU with 100 hidden states. In test time we follow \cite{chen2016bridging} and set the step-size to $\eta = 0.001$. We found SGLD significantly under-performs, so instead we report the performances of two optimizers Adam \citep{kingma2014adam} and Santa taken from \cite{chen2016bridging}. Again these two optimizers use moving average schemes which is out of the scope of our framework, so we mainly compare the meta sampler with SGHMC and leave the others as references.

\begin{figure}[t]
\centering
\subfloat{\includegraphics[width=0.33\textwidth, height=3.5cm]{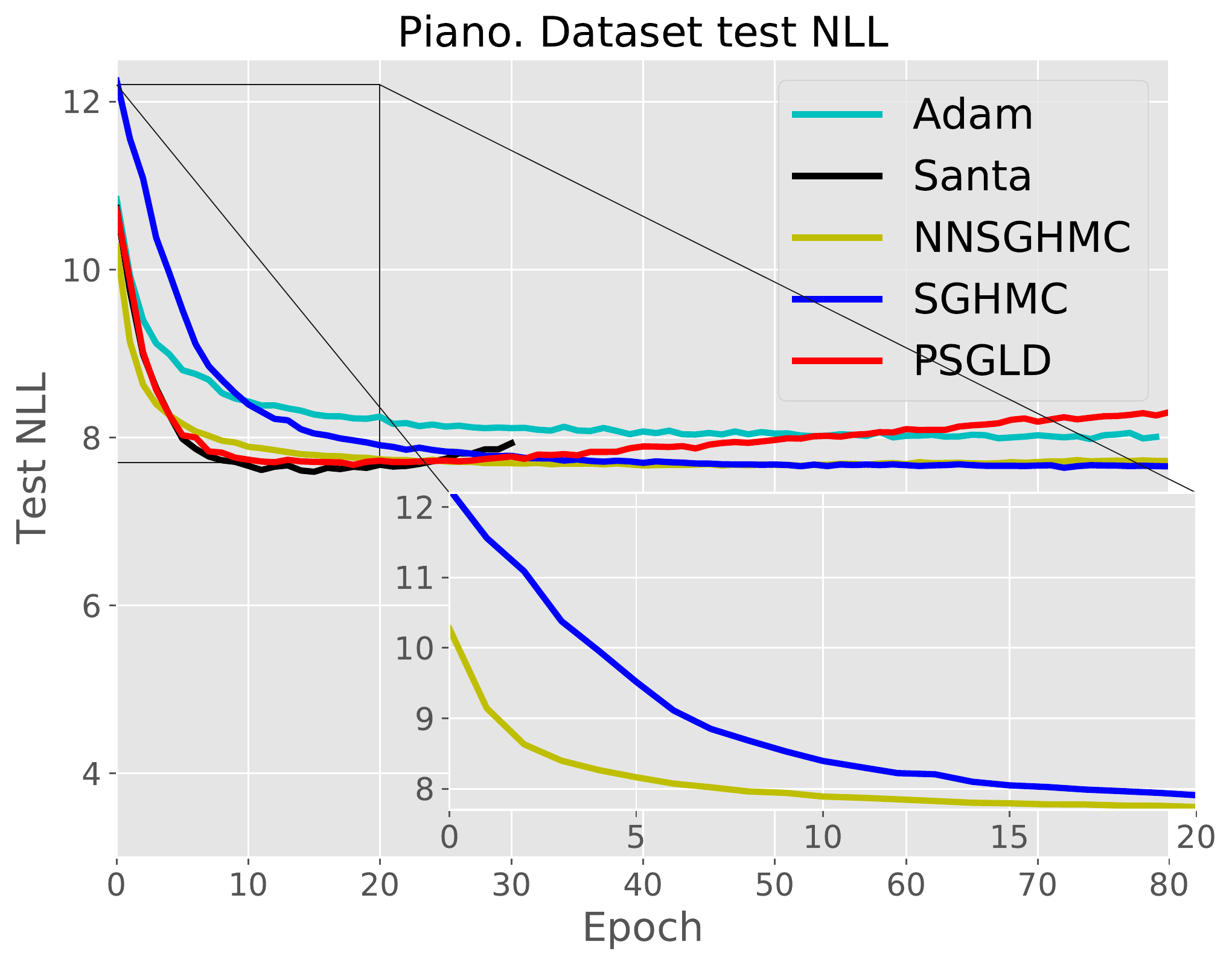}}
\subfloat{\includegraphics[width=0.33\textwidth,height=3.5cm]{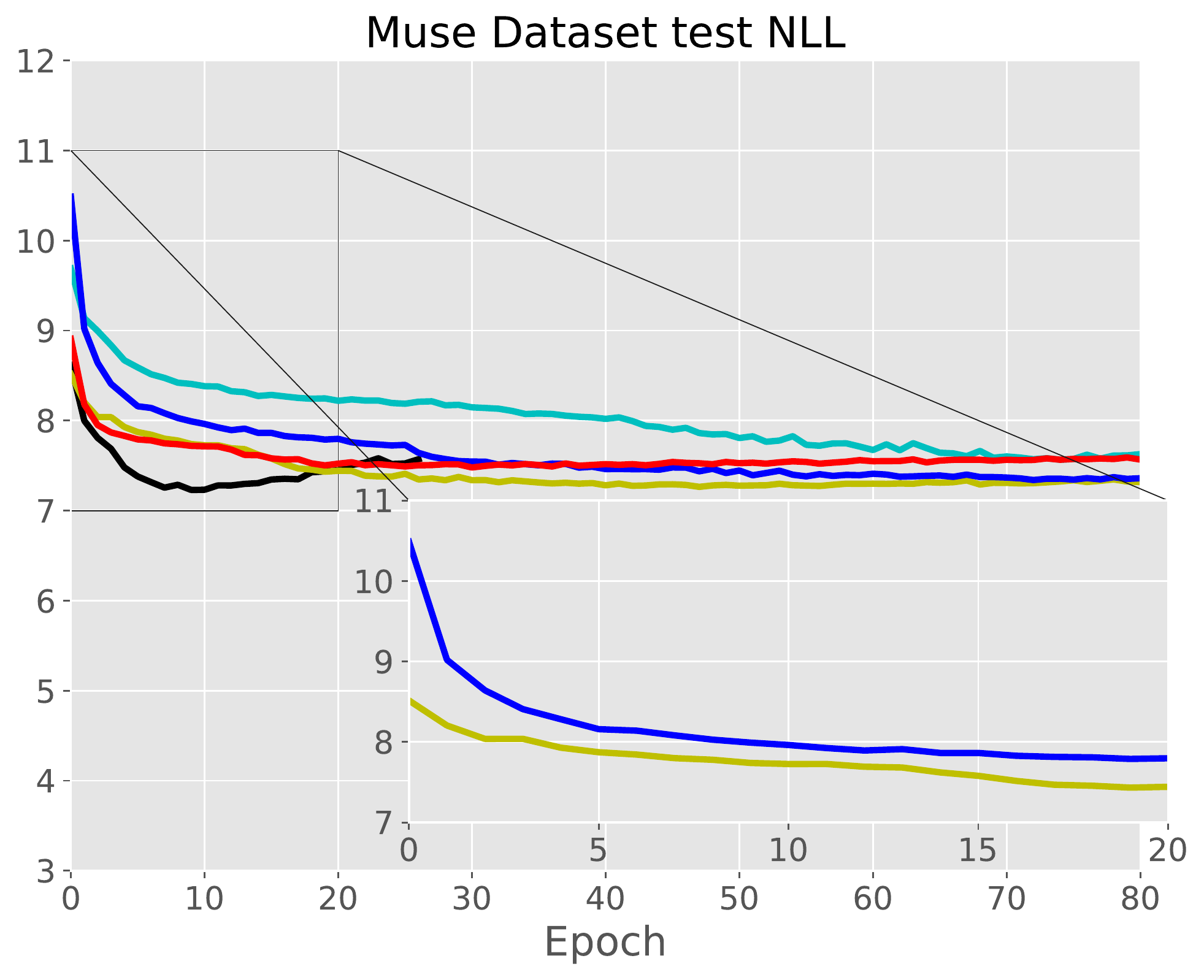}}
\subfloat{\includegraphics[width=0.33\textwidth,height=3.5cm]{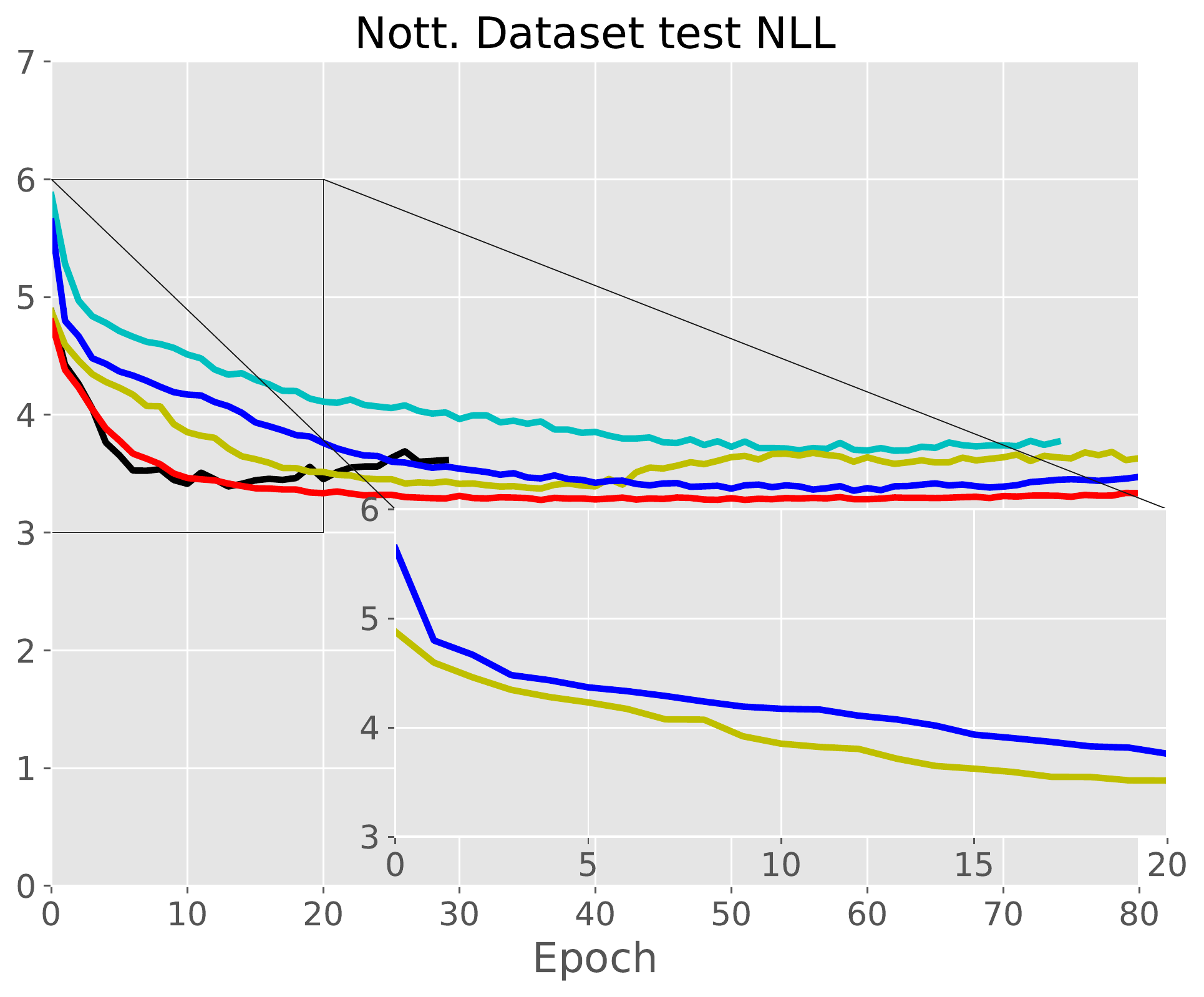}}\\
\hspace{-1.8cm}\subfloat{\includegraphics[width=0.335\textwidth,height=3.5cm]{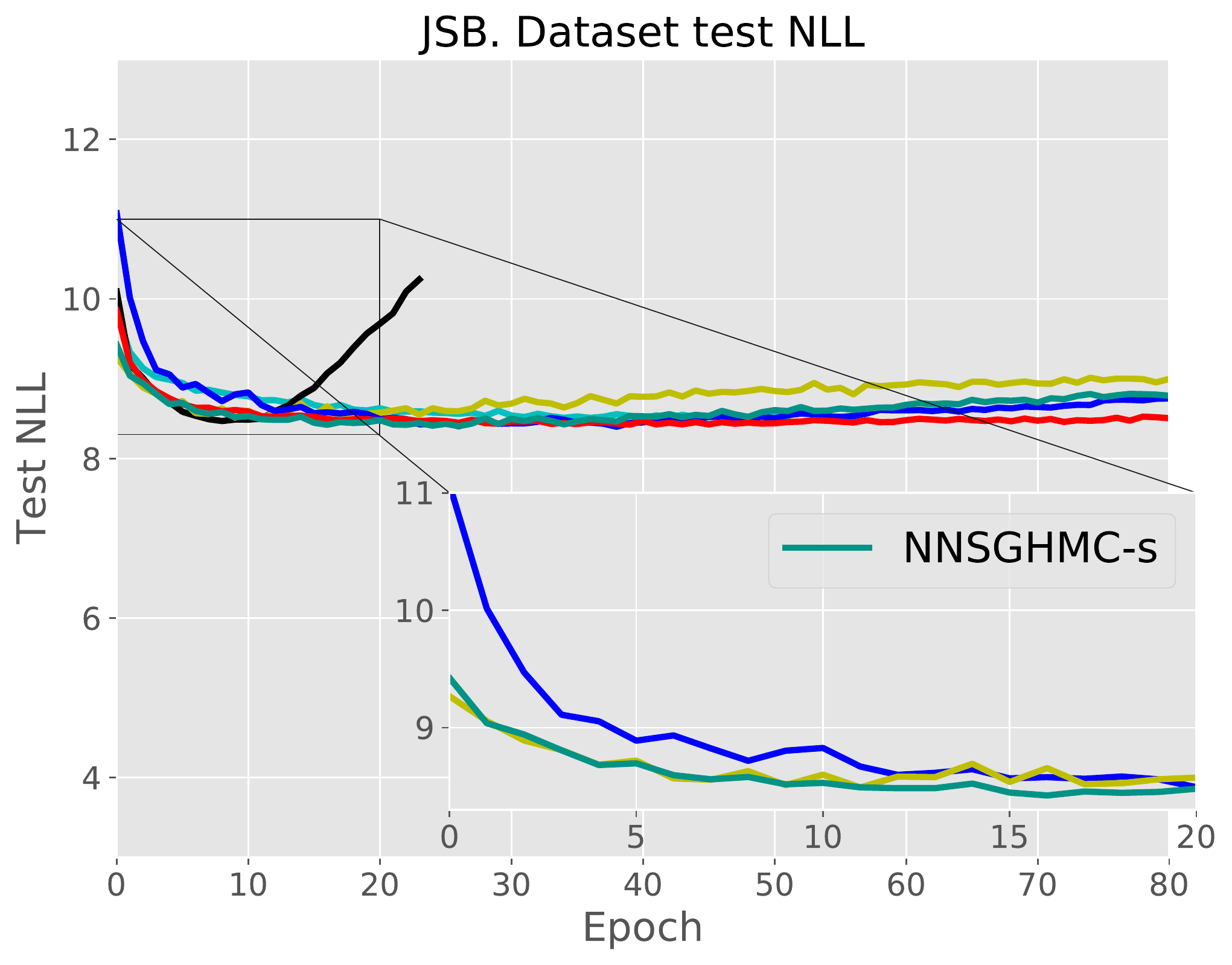}}\hspace{0.3cm}
\subfloat{
\raisebox{4ex}{
\begin{tabular}[b]{ccccc}
\hline
Method&Piano&Muse&Nott&JSB\\
\hline
NNSGHMC&7.66&7.27&3.37&8.49\\
SGHMC&7.65&7.33&3.35&8.40\\
PSGLD&7.67&7.48&3.28&8.42\\
\hline
Santa&7.6&7.2&3.39&8.46\\
Adam&8&7.56&3.7&8.51\\
\hline
\end{tabular}
}}
\caption{Test NLL learning curve (with zoom-in for sampling methods) and the best performance. Santa and Adam results are from \cite{chen2016bridging}}
\label{fig: RNN Curve}
\vspace{-0.1in}
\end{figure}


The meta sampler is tested on the four datasets using 200 unit GRU. For Piano this corresponds to architecture generalization only, and from Figure \ref{fig: RNN Curve} we see that the meta sampler achieves faster convergence compared to SGHMC and achieves similar speed as Santa at early stages. All the samplers achieve best results close to Santa on Piano. The meta sampler successfully generalizes to the other three datasets, demonstrating faster convergence than SGHMC consistently, and better final performance on Muse. 
Interestingly, the meta sampler's final results on Nott and JSB are slightly worse than other samplers. Presumably these two datasets are very different from Muse and Piano, therefore the energy landscape is less similar to the training density (see appendix). Specifically JSB is a dataset with much shorter sequences, in fact SGHMC also exhibits over-fitting but less severe.
Therefore, we further test the meta sampler on JSB without the offset $\beta$ in $\pmb{f}_{\phi_Q}$ to reduce the acceleration (denoted as NNSGHMC-s). Surprisingly, NNSGHMC-s convergences in similar speeds as the original one, but with less amount of over-fitting and better final test NLL \textbf{8.40}.


\section{Conclusions and future work}
We have presented a meta-learning algorithm that can learn an SG-MCMC sampler on simpler tasks and generalizes to more complicated densities in high dimensions. Experiments on both Bayesian MLPs and Bayesian RNNs confirmed strong generalization of the trained sampler to long-time horizon as well as across datasets and network architecture.
Future work will focus on better designs for both the sampler and the meta-learning procedure. For the former, temperature variable augmentation as well as moving average estimation will be explored. For the latter, better loss functions will be proposed for faster training, e.g.~by reducing the unrolling steps of the sampler during training. Finally, the automated design of generic MCMC algorithms that might not be derived from continuous Markov processes remains an open challenge.

\subsection*{Acknowledgements}
We thank Shixiang Gu, Mark Rowland and Cheng Zhang for comments on the manuscript. WG is supported by the CSC-Cambridge Trust Scholarship.

\bibliographystyle{plainnat}
\bibliography{sample}

\clearpage
\appendix
\section{Finite difference approximation for the Gamma vector}
The main computational burden is the gradient computation required by $\pmb{\Gamma}(\pmb{z})$ vector. From the parametrization of $\pmb{Q}$ and $\pmb{D}$ matrix in (\ref{eq: Def of D and Q}), for $\pmb{\theta},\pmb{p}\in \mathbb{R}^D$ we have $\pmb{\Gamma}(\pmb{z}) = [\pmb{\Gamma}_{\pmb{\theta}}, \pmb{\Gamma}_{\pmb{p}}]$. For the first term $\pmb{\Gamma}_{\pmb{\theta}}$, we have
\begin{equation}
\pmb{\Gamma}_{\pmb{\theta},i} =-\nabla_{\pmb{p}}\cdot\pmb{Q}_{i,:} 
=-\frac{\partial f_{\phi_Q,i}}{\partial p_i}.
\label{Appd: Gamma theta}
\end{equation}
Due to the two-stage update of Euler integrator, at time t, we have $f_{\phi_Q,i}^{t-1}=f_{\phi_Q,i}(U(\pmb{\theta}_{t-1}),p_{i}^{t-1})$, $\hat{f}_{\phi_Q,i}^{t-1}=f_{\phi_Q,i}(U(\pmb{\theta}_{t-1}),p_{i}^t)$ and $f_{\phi_D,i}^{t-1}=f_{\phi_D,i}(U(\pmb{\theta}_{t-1}),p_{i}^{t-1},\nabla_{\theta_{i}^{t-1}}U(\pmb{\theta}_{t-1}))$.
Thus a proper finite difference method requires $f_{\phi_Q,i}(U(\pmb{\theta}_t),p_{i}^{t-1})$, which is not exactly the history from the previous time. Therefore we further approximate it using delayed estimate:
\begin{equation}
\frac{\partial f_{\phi_Q,i}^t}{\partial p_i^t} \approx \frac{\hat{f}_{\phi_Q,i}^{t-1}-f_{\phi_Q,i}^{t-1}}{p_{i}^{t}-p_{i}^{t-1}} \quad
\Rightarrow \quad \pmb{\Gamma}_{\pmb{\theta}}^t \approx-\frac{\hat{\pmb{Q}}^{t-1}-\pmb{Q}^{t-1}}{\pmb{p}^t-\pmb{p}^{t-1}}.
\end{equation}
Similarly, the $\pmb{\Gamma}_{\pmb{p}}$ term expands as 
\begin{equation}
\begin{split}
\pmb{\Gamma}_{\pmb{p},i}&=\nabla \cdot [\pmb{D}+\pmb{Q}]_{i,:}\\
&=\frac{\partial f_{\phi_Q,i}}{\partial \theta_i}+\frac{\partial f_{\phi_D,i}}{\partial p_i}+2\alpha f_{\phi_Q,i}\frac{\partial f_{\phi_Q,i}}{\partial p_i}\\
&=\frac{\partial f_{\phi_Q,i}}{\partial U(\pmb{\theta})}\frac{\partial U(\pmb{\theta})}{\partial \theta_i}+\frac{\partial f_{\phi_D,i}}{\partial p_i}+2\alpha f_{\phi_Q,i}\frac{\partial f_{\phi_Q,i}}{\partial p_i}.
\end{split}
\end{equation}
We further approximate $\frac{\partial f_{\phi_Q,i}}{\partial U(\pmb{\theta})}$ by the following
\begin{equation}
\begin{split}
\frac{\partial f_{\phi_Q,i}}{\partial U(\pmb{\theta})}&\approx \frac{f_{\phi_Q,i}^{t}-\hat{f}_{\phi_Q,i}^{t-1}}{U(\pmb{\theta}_t)-U(\pmb{\theta}_{t-1})}
\end{split}
\end{equation}
This only requires the storage of previous $\pmb{Q}$ matrix. However, $\frac{\partial f_{\phi_D,i}}{\partial p_i}$ requires one further forward pass to obtain $\hat{f}_{\phi_D,i}^{t-1}=f_{\phi_D,i}(U(\pmb{\theta}_{t}),p_{i}^{t-1},\nabla_{\theta_i^{t}}U(\pmb{\theta}_t))$, thus, we have 
\begin{equation}
\begin{split}
\frac{\partial f_{\phi_D,i}}{\partial p_i}&\approx \frac{f_{\phi_D,i}^t-\hat{f}_{\phi_D,i}^{t-1}}{p_{i}^t-p_{i}^{t-1}}\\
\Rightarrow\pmb{\Gamma}_{\pmb{p}}&\approx\frac{\pmb{Q}^t-\hat{\pmb{Q}}^{t-1}}{U(\pmb{\theta}_t)-U(\pmb{\theta}_{t-1})}\odot\nabla_{\pmb{\theta}}{U(\pmb{\theta}_t)}+\frac{\pmb{f}_{\phi_D}^t-\hat{\pmb{f}}_{\phi_D}^{t-1}}{\pmb{p}^t-\pmb{p}^{t-1}}+2\alpha \pmb{f}_{\phi_Q}^t\frac{\hat{\pmb{Q}}^{t-1}-\pmb{Q}^{t-1}}{\pmb{p}^t-\pmb{p}^{t-1}}.
\end{split}
\end{equation}
Therefore the proposed finite difference method only requires one more forward passes to compute $\hat{\pmb{f}}_{\phi_D}^{t-1}$ and instead, save 3 back-propagations. As back-propagation is typically more expensive than forward pass, our approach reduces running time drastically, especially when the sampler are applied to large neural network.

\paragraph{Time complexity figures} 
Every SG-MCMC method (including the meta sampler) requires $\nabla_{\pmb{\theta}}U(\pmb{\theta})$. The main burden is the forward pass and back-propagation through the $\pmb{D}$ and $\pmb{Q}$ matrix, where the latter one has been replaced by the proposed finite difference scheme. The time complexity is $O(HD)$ for both forward pass and finite difference with $H$ the number of hidden units in the neural network of the meta sampler. Parallel computation with GPUs improves real-time speed, indeed in our MNIST experiment the meta sampler spends roughly 1.5x time when compared with SGHMC.

During meta sampler training, the Stein gradient estimator requires the kernel matrix inversion which is $O(K^3)$ for cross-chain training. In practice, we only run a few parallel Markov chains $K=20 \sim 50$, thus, this will not incur huge computation cost. For in-chain loss the computation can also be reduced with proper thinning schemes.

\section{Training details}
\begin{figure}
\subfloat{\includegraphics[width=0.45\textwidth,height=4cm]{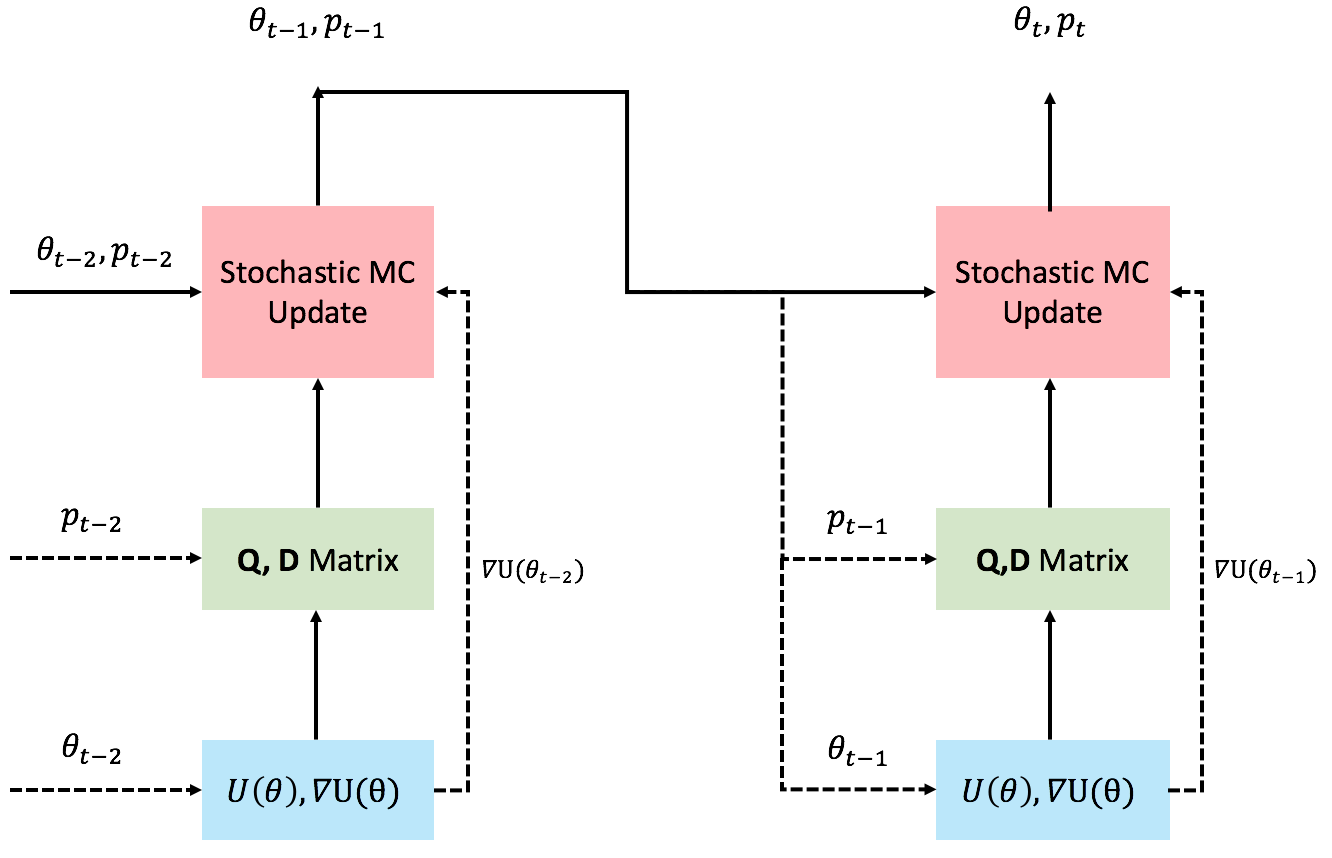}}\hfill
\subfloat{\includegraphics[width=0.55\textwidth,height=4cm]{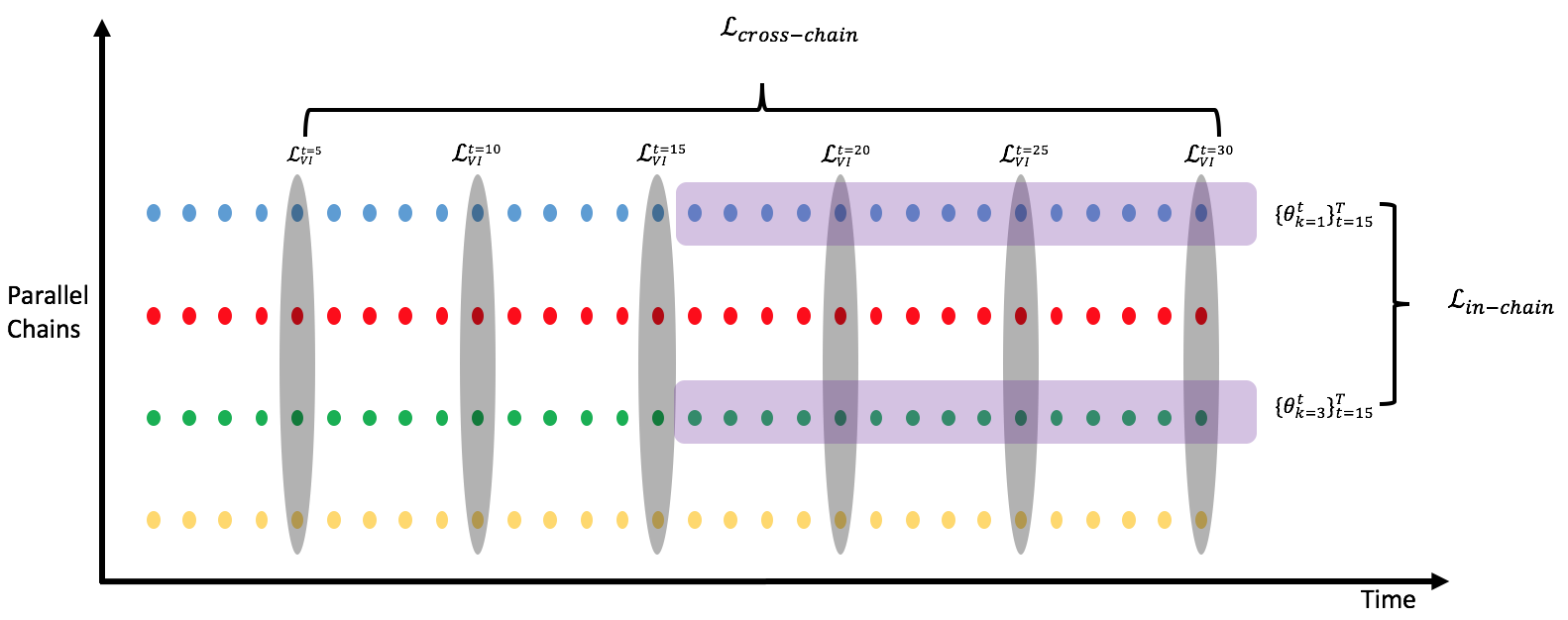}}
\caption{(Left) The unrolled scheme of the meta sampler updates. Stop gradient operations are applied to the \textit{dashed} arrows. (Right) A visualization of cross-chain in-chain training. The grey area represents samples across multiple chains, and we compute the cross chain loss for every 5 time steps. The purple area indicates the samples taken across time with sub-sampled chains 1 and 3. In this visualization the initial 15 samples are discarded for burn-in, and the thinning length is $\tau = 1$ (effectively no thinning).}
\label{fig: Stop Gradient and Cross Chain In Chain Loss}
\end{figure}

We visualize on the left panel of Figure \ref{fig: Stop Gradient and Cross Chain In Chain Loss} the unrolled computation scheme. We apply truncated back-propagate through time (BPTT) to train the sampler. Specifically, we manually stop the gradient flow through the input of $\pmb{D}$ and $\pmb{Q}$ matrix to avoid computing higher order gradients. 
 
We also illustrate cross-chain in-chain training on the right panel of Figure \ref{fig: Stop Gradient and Cross Chain In Chain Loss}. Cross-chain training encourages both fast convergence and low bias, provided that the samples are taken from parallel chains. 
On the other hand, in-chain training encourages sample diversity inside a chain. In practice, we might consider thinning the chains when performing in-chain training. Empirically this improves the Stein gradient estimator's accuracy as the samples are spread out. Computationally, this also prevents inverting big matrices for the Stein gradient estimator, and reduces the number of back-propagation operations. 
Another trick we applied is parallel chain sub-sampling: if all the chains are used, then there is less encouragement of singe chain mixing, since the parallel chain samples can be diverse enough already to give reasonable gradient estimate. 

\section{Input pre-processing}
One potential challenge is that for different tasks and problem dimensions, the energy function, momentum and energy gradient can have very different scales and magnitudes. This affects the meta sampler's generalization, for example, if training and test densities have completely different energy scales, then the meta sampler is likely to produce wrong strategies. This is especially the case when the meta sampler is generalized to much bigger networks or to very different datasets.

To mediate this issue, we propose to pre-process the inputs to both $\pmb{f}_{\phi_D}$ and $\pmb{f}_{\phi_Q}$ networks to make it at similar scale as those in training task. Recall that the energy function is $U(\pmb{\theta})=-\sum_{n=1}^N{\log p(\pmb{y}_n|\pmb{x}_n,\pmb{\theta})-\log p(\pmb{\theta})}$ where the prior $\log p(\pmb{\theta})$ is often isotropic Gaussian distribution. Thus the energy function scale linearly w.r.t~ both the dimensionality of $\pmb{\theta}$ and the total number of observations $N$. Often the energy function is further approximated using mini-batches of $M$ datapoints. Putting them together, we propose pre-processing the energy as 
\begin{equation}
\overbar{U(\pmb{\theta})}=\frac{1}{M}\sum_{m=1}^{M}{\log p(\pmb{y}_m|\pmb{x}_m,\pmb{\theta})}+\frac{D_{\text{train}}}{ND_{\text{test}}}\log p(\pmb{\theta})
\label{eq: energy scaling}
\end{equation}
where $D_{\text{train}}$ and $D_{\text{test}}$ are the dimensionality of $\pmb{\theta}$ in the training task and the test task, respectively. 
Importantly, for RNNs $N$ represents the total sequence length, namely $N=\sum_{n=1}^{N_{data}}{T_n}$, where $N_{data}$ is the total number of sequences and $T_n$ is the sequence length for a datum $\pmb{x}_n$. We also define $M$ accordingly. The momentum and energy gradient magnitudes are estimated by simulating a randomly initialized meta sampler for short iterations. With these statistics we normalize both the momentum and the energy gradient to have roughly zero mean and unit variance.

\section{Experiment Setup}
\subsection{Toy Example}
We train our meta sampler on a 10D uncorrelated Gaussian with mean $(3, ..., 3)$ and randomly generated covariance matrix. We do not set any offset and additional frictions, i.e. $\alpha=0$ and $\beta=0$. The noise estimation matrix $\tilde{\pmb{B}}$ are set to be $0$ for both meta sampler and SGHMC. To mimic stochastic gradient, we manually inject Gaussian noise with zero mean and unit variance into $\nabla_{\pmb{\theta}}\tilde{U}(\pmb{\theta})=\nabla_{\pmb{\theta}}U(\pmb{\theta})+\pmb{\epsilon}, \pmb{\epsilon}\sim \mathcal{N}(\pmb{0},\pmb{I})$. The functions $\pmb{f}_{\phi_D}$ and $\pmb{f}_{\phi_Q}$ are represented by 1-hidden-layer MLPs with 40 hidden units. For training task, the meta sampler step size is 0.01. The initial positions are drawn from $\text{Uniform}([0, 6]^D)$. 
We train our sampler for 100 epochs and each epochs consists 4 x 100 steps. For every 100 steps, we updates our $\pmb{Q}$ and $\pmb{D}$ using Adam optimizer with learning rate 0.0005. Then we continue the updated sampler with last position and momentum until 4 sub-epochs are finished. We re-initialize the momentum and position. 
We use both cross-chain and in-chain losses. The Stein Gradient estimator uses RBF kernel with bandwidth chosen to be 0.5 times the median-heuristic estimated value. We unroll the Markov Chain for 20 steps before we manually stop the gradient. For cross-chain training, we take sampler across chain for each 2 time steps. For in-Chain, we discard initial 50 points for burn-in and sub-sample the chain with batch size 5. We thin the samples for every 3 steps. For both training and evaluation, we run 50 parallel Markov Chains.

The test task is to draw samples from a 20D correlated Gaussian with with mean $(3, ..., 3)$ and randomly generated covariance matrix. The step size is 0.025 for both meta sampler and SGHMC. To stabilize the meta sampler we also clamp the output values of $\pmb{f}_{\phi_Q}$ within $[-5,5]$. The friction matrix for SGHMC is selected as $\pmb{I}$.

\subsection{Bayesian MLP MNIST}
In MNIST experiment, we apply input pre-processing on energy function as in (\ref{eq: energy scaling}) and scale energy gradient by 70. Also, we scale up $\pmb{f}_{\phi_D}$ by 50 to account for sum of stochastic noise. The offset $\alpha$ is selected as $\frac{0.01}{\eta}$ as suggested by \cite{chen2014stochastic}, where $\eta = \sqrt{\frac{lr}{N}}$ with $lr$ the per-batch learning rate.
We also turn off the off-set and noise estimation, i.e. $\beta =0$ and $\tilde{\pmb{B}}=0$. We run 20 parallel chains for both training and evaluation. We only adopt the cross chain training with thinning samplers of 5 times step. We also use the finite difference technique during evaluation to speed-up computations. 

\subsubsection{Architecture Generalization}
We train the meta sampler on a smaller BNN with architecture 784-20-10 and ReLU activation function, then test it on a larger one with architecture 784-40-40-10. In both cases the batch size is 500 following \cite{chen2014stochastic}. Both $\pmb{f}_{\phi_D}$ and $\pmb{f}_{\phi_Q}$ are parameterized by 1-hidden-layer MLPs with 10 units. The per-batch learning rate is 0.007. We train the sampler for 100 epochs and each one consists of 7 sub-epochs. For each sub-epoch, we run the sampler for 100 steps. We re-initialize $\pmb{\theta}$ and momentum after each epoch. To stabilize the meta sampler in evaluation, we first run the meta sampler with small per-batch learning rate $0.0085$ for 3 data epochs and clamp the $\pmb{Q}$ values. After, we increase the per-batch learning rate to $0.018$ with clipped $\pmb{f}_{\phi_Q}$. The learning rate for SGHMC is $0.01$ for all times. For SGLD and PSGLD, they are $0.2$ and $1.4\times 10^{-3}$ respectively. These step-sizes are tuned on MNIST validation data.

\subsubsection{Activation function generalization}
We modify the test network's activation function to \textbf{sigmoid}. We use almost the same settings as in network generalization tests, except that the per-batch learning rates are tuned again on validation data. For the meta sampler and SGHMC, they are $0.18$ and $0.15$. For SGLD and PSGLD, they are $1$ and $1.3\times 10^{-2}$.

\subsection{Dataset Generalization}
We train the meta sampler on ReLU network with architecture 784-20-5 to classify images 0-4, and test the sampler on ReLU network 784-40-40-5 to classify images 5-9. The settings are mostly the same as in network architecture generalization for both training and evaluation. One exception is again the per-batch learning rate for PSGLD, which is tuned as $1.3\times 10^{-3}$. Note that even though we use the same per-batch learning rate as before, the discretization step-size is now different due to smaller training dataset, thus, $\alpha$ will be automatically adjusted accordingly.

\subsection{Bayesian RNN}
The \emph{Piano} data is selected as the training task, which is further split into training, validation and test subsets. We use batch-size 1, meaning that the energy and the gradient are estimated on a single sequence. The meta sampler uses similar neural network architectures as in MNIST tests. The training and evaluation per-batch learning rate for all the samplers is set to be 0.001 following \cite{chen2016bridging}. 
We train the meta sampler for 40 epochs with 7 sub-epochs with only cross chain loss. Each sub-epochs consists 70 iterations. We scale the $\pmb{D}$ output by 20 and set $\alpha=\frac{0.002}{\eta}$, where $\eta$ is defined in the same way as before. We use zero offset during training, i.e. $\beta=0$. We apply input pre-processing for both $\pmb{f}_{\phi_D}$ and $\pmb{f}_{\phi_Q}$. To prevent divergence of the meta sampler at early training stage. We also set the constant of $c=100$ to the  $f_{\phi_D}$. For dataset generalization, we tune the off-set value based on Piano validation set and transfer the tuned setting $\beta=-1.5$ to the other three datasets. For Piano architecture generalization, we do not tune any hyper-parameters including $\beta$ and use exactly same settings as training. Exact gradient is used in RNN experiments instead of computing finite differences. 

\section{RNN dataset description}
We list some data statistics in Table \ref{tab: RNN dataset statistics} which roughly indicates the similarity between datasets. 
\begin{table}
\centering
\caption{The basic statistics for 4 RNN datasets, bold figure represents large difference compared to others. \textit{Size} is the number of data point. \textit{Avg. Time} is the averaged sequence and \textit{Energy scale} is the rough scale of the train NLL when sampler converges.}
\begin{tabular}{c|cccc}
\hline
&Piano&Muse&Nott&JSB\\
\hline
Size:train&\textbf{87}&524&694&229\\
Size:test&25&124&170&77\\
Avg. Time:train&872&467&254&\textbf{60}\\
Avg. Time:test&761&518&261&\textbf{61}\\
Energy scale:train&$\approx 7.2$&$\approx 7$&$\approx\pmb{2.5}$&$\approx 7.8$\\
\hline
\end{tabular}
\label{tab: RNN dataset statistics}
\end{table}
Piano dataset is the smallest in terms of data number, however, the averaged sequence length is the largest. Muse dataset is similar to Piano in sequence length and energy scale but much larger in terms of data number. On the other hand, Nott dataset has very different energy scale compared to the other three. This potentially makes the generalization much harder due to inconsistent energy scale fed into $\pmb{f}_{\phi_Q}$ and $\pmb{f}_{\phi_D}$. For JSB, we notice a very short sequence length on average, therefore the GRU model is more likely to over-fit. Indeed, some algorithms exhibits significant over-fitting behavior on JSB dataset compared to other data (Santa is particularly severe).

\section{Additional Plots}
\subsection{Short run comparison}
We also run the samplers using the same settings as in MNIST experiments for a short period of time (500 iterations). We also compare to other optimization methods including Momentum SGD (SGD-M) and Adam. We use the same per-batch learning rate for SGD-M and SGHMC as in MNIST experiment. For Adam, we use 0.002 for ReLU and 0.01 for Sigmoid network. 

\begin{figure}
\centering
\includegraphics[width=1\textwidth,height=0.65\textwidth]{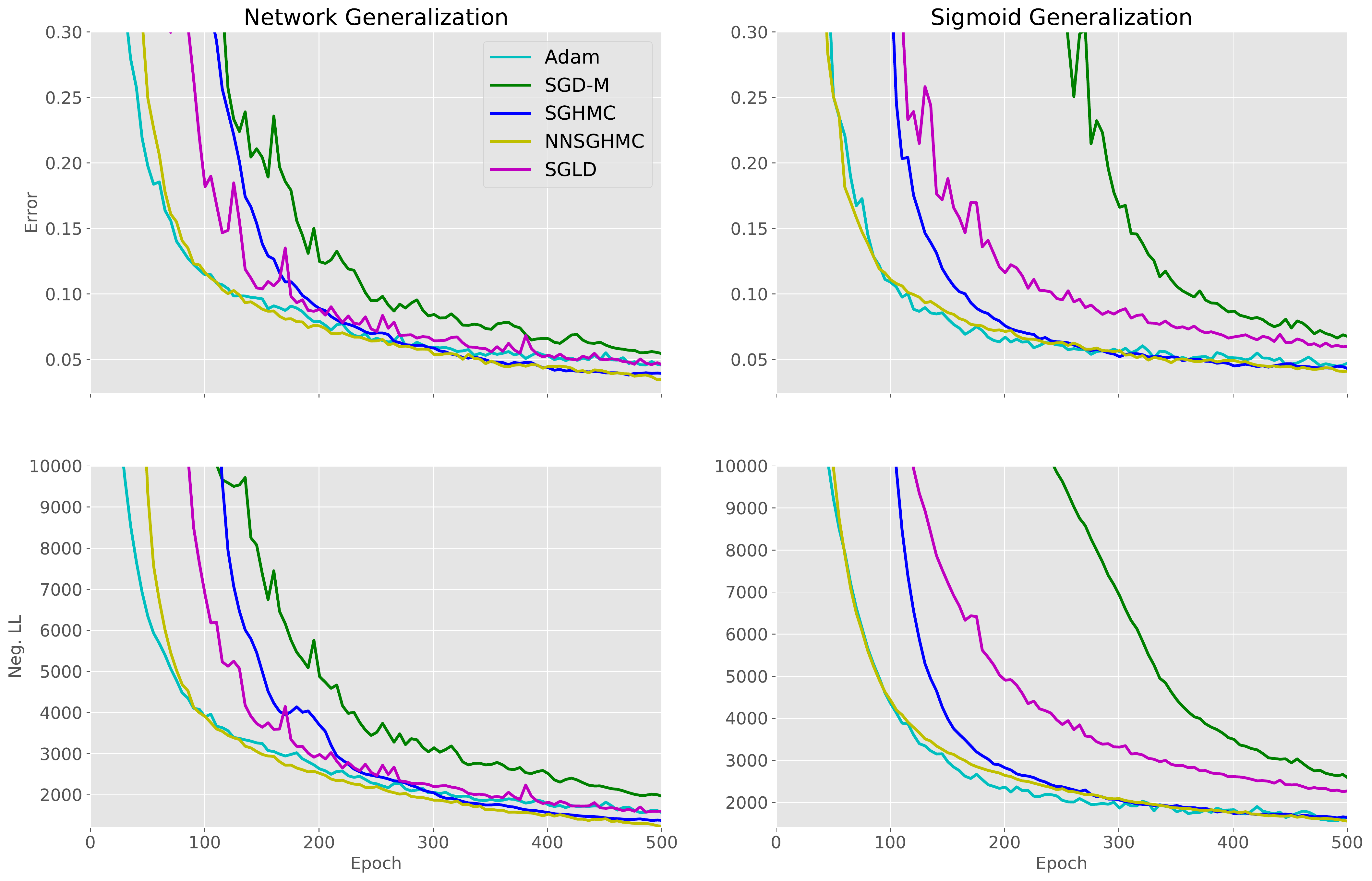}
\caption{We only test the \textit{Network Generalization} and \textit{Activation function generalization}. The \textbf{upper} part indicates the test error plot and \textbf{lower} part are the negative test LL curve}
\label{fig:Additional Short Run}
\end{figure}

\begin{figure}[!t]
\centering
\includegraphics[width=0.7\textwidth,height=0.6\textwidth]{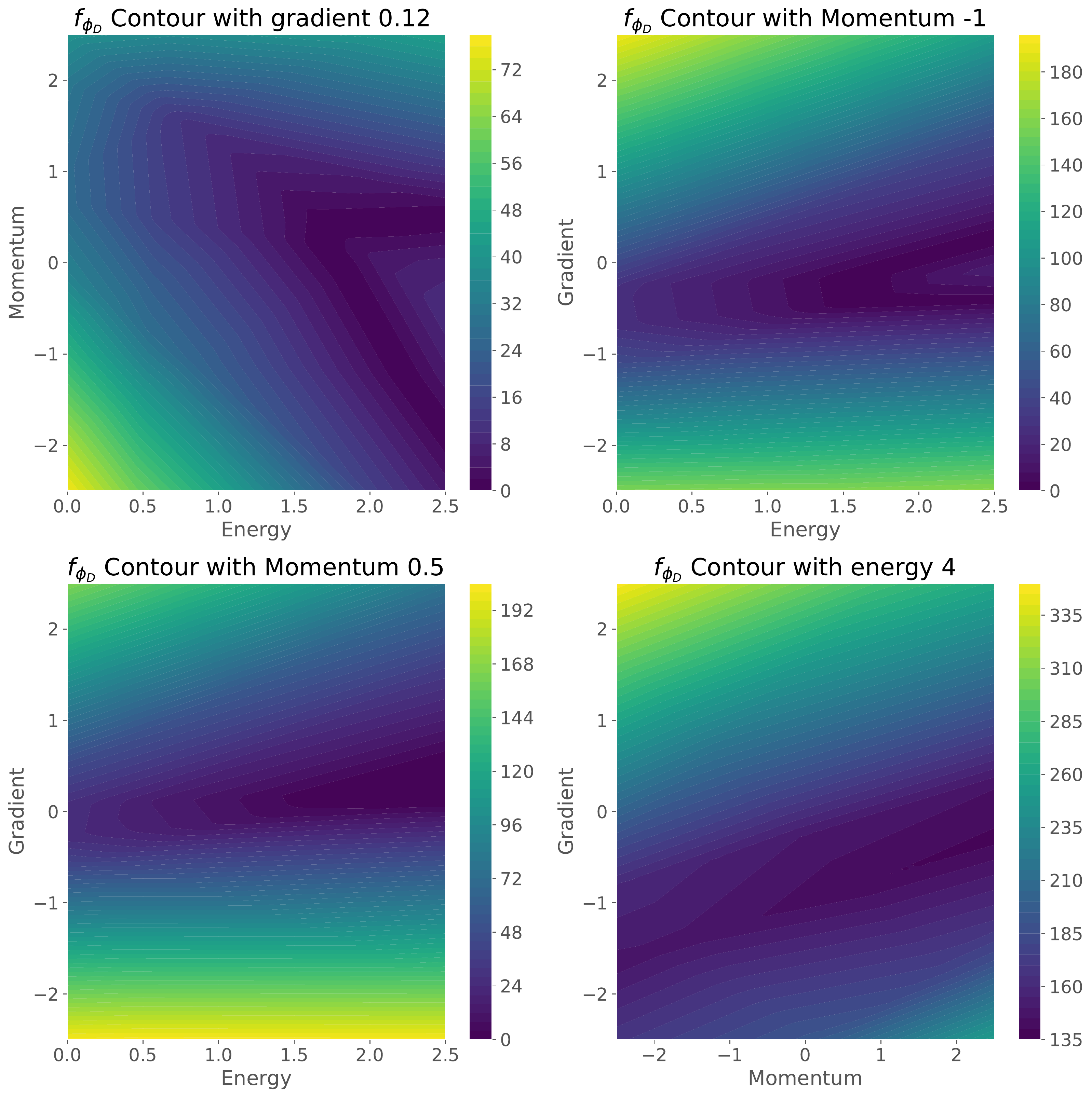}
\caption{The contour plots of $f_{\phi_D}$ for other input values.}
\label{fig: Contour left}
\end{figure}

The results are shown in Figure \ref{fig:Additional Short Run}.
Meta sampler and Adam achieves the fastest convergence speed. 
%
%
This again confirms the faster convergence of the meta sampler especially at initial stages. 
We also provide additional contour plots (Figure \ref{fig: Contour left}) to demonstrate the strategy learned by $\pmb{f}_{\phi_D}$ for reference.

\end{document}